\documentclass[runningheads]{llncs}
\usepackage{graphicx}
\usepackage{amsmath,amssymb} % define this before the line numbering.
\usepackage{color}
\usepackage[noend,linesnumbered,figure,boxed]{algorithm2e}

\begin{document}
%\renewcommand\thelinenumber{\color[rgb]{0.2,0.5,0.8}\normalfont\sffamily\scriptsize\arabic{linenumber}\color[rgb]{0,0,0}}
%\renewcommand\makeLineNumber {\hss\thelinenumber\ \hspace{6mm} \rlap{\hskip\textwidth\ \hspace{6.5mm}\thelinenumber}}
%\linenumbers
\pagestyle{headings}
\mainmatter
%\def\ECCV12SubNumber{***}  % Insert your submission number here

%%%%%%%%% TITLE
\title{Contour Completion Around a Fixation Point}

%\titlerunning{ECCV-12 submission ID \ECCV12SubNumber}

%\authorrunning{ECCV-12 submission ID \ECCV12SubNumber}

%\author{Anonymous ECCV submission}
%\institute{Paper ID \ECCV12SubNumber}
\author{Toshiro Kubota}
\institute{
Susquehanna University\\
Selinsgrove, PA, USA {\tt\small kubota@susqu.edu}}

\maketitle
% \thispagestyle{empty}

%%%%%%%%% ABSTRACT
\begin{abstract}
The paper presents two edge grouping algorithms for finding a
closed contour starting from a particular edge point and
enclosing a fixation point. Both algorithms search a shortest
simple cycle in \textit{an angularly ordered graph} derived
from an edge image where a vertex is an end point of a contour
fragment and an undirected arc is drawn between a pair of
end-points whose visual angle from the fixation point is less
than a threshold value, which is set to $\pi/2$ in our
experiments. The first algorithm restricts the search space by
disregarding arcs that cross the line extending from the
fixation point to the starting point. The second algorithm
improves the solution of the first algorithm in a greedy
manner. The algorithms were tested with a large number of
natural images with manually placed fixation and starting
points. The results are promising.
\end{abstract}

%%%%%%%%% BODY TEXT
\section{Introduction}
Edge grouping or contour integration is an important step for
many computer vision applications. Various methods have been
proposed in the past, which have shown effectiveness in many
controlled situations. However, the algorithms tend to break
down as the amount of noise and clutters increases in the
image. Another issue present in many existing contour
integration algorithms is that it does not offer an intuitive
control of a region of interest and a scale of the object;
algorithms offer a closed contour that is deemed best in terms
of some criteria imposed uniformly to the entire image.
However, the most interesting or salient contour is highly
dependent on the problem and focus of the user/system. Most
algorithms do not provide a simple way to incorporate such
situation dependent options.

In \cite{Kubota:POCV2010}, an algorithm to find a closed
contour that encloses a fixation point and penetrates a
selected edge point is presented. By locating the fixation
point in the area of interest and the edge point on a salient
edge, the algorithm could extract different objects in the
image while implicitly specifying the scale of interest. It
employs dynamic programming to find a shortest path from the
source to the target, and is highly efficient. A serious
shortcoming is that it can only extract a star shape where
every contour point of the shape is visible from the focal
point without being occluded by another part of the shape.

The main objective of this paper is to alleviate the
shortcoming of \cite{Kubota:POCV2010} so that the algorithm is
applicable to more general shapes. Although color information
is rich and essential for our perception and can work jointly
with contour information to enhance segmentation
algorithms\cite{Arbelaez:PAMI2011}\cite{Mobahi:IJCV2011}, we
concentrate our study on edge information only. The reason is
three-fold. First, our vision system can extract a great deal
of information from an edge map with the capability far
exceeding that of the current state-of-the-art edge grouping
algorithms. Thus, we can still improve the performance of
contour integration using edge information only. Once we
achieve our goal, we can further improve the performance by
incorporating the color information to the algorithm.
%Second, edge information can be coded more compactly than color
%information, in general. Thus, an edge-based technique can be
%more advantageous in situations where the communication
%bandwidth is low.
Second, we can associate our studies with various psycho-visual
ones, which often study edges and colors separately. Third, we
can make fair comparisons with many other edge grouping
algorithms, in particular the algorithm of
\cite{Kubota:POCV2010}.

We cast the contour integration problem to a graph search
problem, and ask the following question: find a shortest cycle
that starts at a chosen vertex and encloses a chosen fixation
point. The algorithm of \cite{Kubota:POCV2010} finds a solution
in a restricted space of star shapes. In this paper, we present
two algorithms that can handle larger classes of shapes. The
formulation of arc weights is crucial to graph based edge
grouping algorithms. To align  the current study to that of
\cite{Kubota:POCV2010}, we keep the arc weight formulation
simple without using any tangential and curvature information.
%We believe that the performance of our algorithms can be
%improved by employing more complex formulation of arc weights,
%which we will explore in the future.

\section{Related Works}
In the past, various contour integration algorithms have been
proposed. Many of them formulated the problem as a graph based
one and derived a solution via efficient graph search
algorithms. A graph consists of a set of \textit{vertices} and
a set of \textit{arcs}. (We use 'arc' instead of 'edge' to
avoid confusion with edges of an image.) A sequence of
connected edge pixels (\textit{edgels}) is called \textit{a
contour fragment} in this paper.

In \cite{elder:curve}, the problem is formulated as a shortest
path problem with arc weights encoding tangential information
of contour fragments and coloring information surrounding the
fragments. From a salient contour fragment, Dijkstra's
algorithm is applied to find a shortest cycle. Since the path
cost based on their arc weights increases with the length of
the contour, it tends to extract a short closed contour.

In \cite{Mahamud:PO}, stochastic completion fields proposed by
\cite{Williams:NECO97a} are used to derive transition
probability between a pair of contour fragments, and the
saliency of the transition is derived using the eigenvector of
the transition matrix corresponding to the largest eigenvalue.
Via the transition saliency, a sparse directed graph is
constructed and a strongly connected component with the most
salient arc was extracted as the most salient closed contour.
Since the transition saliency is a global construct in their
formulation, suppression of the arcs in the most salient closed
contour was needed to expose the second salient one.

In \cite{Wang:RatioContour}, elastica was used to define arc
weights and a minimum perfect matching was used to derive a
closed contour with the smallest ratio of the total arc weights
and the length of the contour. By using the ratio form, the
algorithm avoided favoring short contours. The algorithm,
however, was restricted to provide only the most salient closed
contour. As in \cite{Mahamud:PO}, extraction of secondary
salient contours required suppression of arcs in the most
salient solution.

In \cite{Stahl:PAMI2008}, symmetry information aided the
grouping process. Symmetry is often a strong cue for man-made
objects. However, it is difficult to incorporate the
information into arc weights as it is a non-local property. (In
contrast, proximity and continuity are local properties.) The
authors devised an ingenious way to incorporate symmetry by
introducing symmetric trapezoids derived from a pair of contour
fragments as grouping tokens. This work used the same graph
search method of \cite{Wang:RatioContour}. Hence, it
experienced the same issue in extracting secondary contours as
in \cite{Wang:RatioContour} and \cite{Mahamud:PO}.
%Since symmetry may not as strong cue in natural objects as in
%man-made ones, the method may be less effective for outdoor
%scenes than indoor ones.

In these algorithms, a saliency condition is encoded in the arc
weights of the graph and is fixed. However, the condition is
often dependent on the goal and a focus of the system. A shape
deemed most salient in one application may not be the most
desired one in another. Even within the same application, the
saliency criteria can change dynamically, for example, during
parsing of the scene for navigation. Many existing algorithms
including those mentioned above do not provide mechanisms to
change the focus or the saliency criteria in an intuitive
manner. Compounded with the fixed arc weight issue is that the
algorithms only provide a single optimal solution. When the
system demands for multiple solutions, the algorithms need to
suppress extracted arcs. This approach has an undesired
consequence that subsequent solutions cannot share arcs
selected in previous solutions.

In \cite{Kubota:POCV2010}, a fixation point and a starting
point were introduced and the algorithm derived a cyclic path
starting from the chosen starting point and surrounding the
fixation point. By incorporating the starting point, a user has
precise control of where the solution begins. With the fixation
point together with the starting point, the user also has
control of the size of the object. The paper also proposed
simple procedures to automate placement of starting positions
and to extract multiple contours around the fixation point in
succession. With the algorithm of \cite{elder:curve}, a user
can select a starting position of the closed contour. However,
the search is still driven toward a short contour. Since our
algorithms are based closely with the algorithm of
\cite{Kubota:POCV2010}, we describe it in more detail below.

The first step of the algorithm is to divide the $2\pi$ field
of view from the fixation point into an equally spaced set of
$M$ bins, and place each edgel in one of the bins except the
edgel at the starting position. Two additional bins are
attached before and after the $M$ bins and the starting edgel
is placed in both bins. Edgels that were in the same bin and
also are 8-connected in the edge image were aggregated into a
super-edgel. For a super-edgel $x$ in $i$th bin, allowed
transition was restricted to super-edgels in bins from $i$ to
$i+m$ where $m\ge 0$ is a parameter that controls the size of a
gap allowed in the solution. Hence the maximum gap allowed is
capped at $(m/M)2\pi$. Using this set-up, the algorithm finds a
shortest path from the only vertex in the first bin (the
starting edgel) to the only vertex in the last bin (the
duplicate of the starting edgel).
%For illustration of this algorithm, see Figure
%\ref{fig:POCVIllustration} where (a) is a simple edge image
%that is divided into 8 bins around a fixation point shown in a
%solid disk, and (b) is a polar representation of the edge
%image, which is closely associated with the dynamic programming
%table.
The major limitation of the algorithm is that it can handle
only star-shaped contours where each contour point in the shape
is visible from the fixation point.

The work of \cite{Mishra:iCCV2009} also considers a fixation
point as a parameter to the segmentation algorithm. It
transforms the edge image into the polar coordinate and uses
graph cut to separate inside/outside regions with respect to
the fixation point. A graph is constructed by connecting four
neighbors in the image grid of the polar domain and assigning
weights encoding dissimilarity measures of the pixels.

%\begin{figure}
%\centering
%\includegraphics[width=3.in]{POCVIllustration.eps}
%\caption{Formulation of an undirectional graph from an edge image. (a)
%is an edge image Locations of a fixation point and and an interesting
%point are given. (b) shows a tabulated representation of the edge image. }
%\label{fig:POCVIllustration}
%\end{figure}

%Recently, various methods that incorporate both contour and
%color information for segmentation of natural images have been
%proposed. \cite{Arbelaez:PAMI2011}\cite{Mobahi:IJCV2011}. These
%studies demonstrate that contour and region information can
%complement each other to enhance the segmentation performance
%in carefully designed algorithms. In the future, we will
%incorporate color information into the proposed framework via
%either arc weight formulation that takes the color information
%into account or a fill-in process that runs in parallel with
%the contour process and jointly improve the segmentation
%process.

\section{Formulation}
We take the following preparatory steps on a given digital
(gray scale) image to obtain our graph representation. First,
we apply Canny edge detector. Junctions and end-points of
connected edgels are detected and contour fragments are formed
by tracing every end-point to either another end-point or a
junction. We impose the maximum length of $L$ pixels to contour
fragments. Hence, each contour fragment whose length exceed $L$
is split into multiple fragments. This splitting step is to
reduce the risk of merging two contour fragments that belong to
different objects in the image, and is not a critical part of
the overall algorithm. We chose $L=10$ in our experiments.

An undirected graph ($G$) is constructed by treating each
end-point of each contour fragment as a vertex and connecting
every pair of vertices with a undirected arc whose weight is
computed as follows. Let $\textbf{u}$ and $\textbf{v}$ be the
two vertices of the edge. If $\textbf{u}$ and $\textbf{v}$ are
on the same contour fragment, the weight is set to 0. If they
are from a different contour fragment, the weight is set to the
square of the Euclidean distance between two end-points
represented by $\textbf{u}$ and $\textbf{v}$. We choose to
square the distance so that a contour comprised of a large gap
is penalized more than a contour comprised of many small gaps.
We used neither differential information such as tangent and
curvature nor color information, to keep the preprocessing
stage as elementally as possible. This also allows fair
comparison with the results of \cite{Kubota:POCV2010}, in which
the cost measure was similarly defined. Two inputs are
provided: a fixation point (\textbf{o}) and an interesting
point in the image. Given the interesting point, we find the
closest edgel, split the fragment there, and designate one of
resulting endpoints as the starting vertex. We use \textbf{s}
to denote the starting vertex.

Now, we can state the problem as follows. Given $G$,
\textbf{o}, and \textbf{s}, find a shortest simple cycle
starting from \textbf{s} in $G$ that encloses \textbf{o}. We
need to be precise about the meaning of enclosure. We say a
cycle is a $\theta$-enclosure of \textbf{o}, if every gap in
the cycle has its visual angle less than $\theta$. We require
$\theta\le\pi$ so that the fixation point always lies
physically inside the $\theta$-enclosure when gaps are
connected by straight line segments.

We can safely remove arcs whose visual angles are not less than
$\theta$ as these arcs cannot be a part of a $\theta$-enclosing
cycle. We call $G$ obtained after the arc removal
\textit{angularly ordered graph}.

\begin{figure}
\centering
\includegraphics[width=4.in]{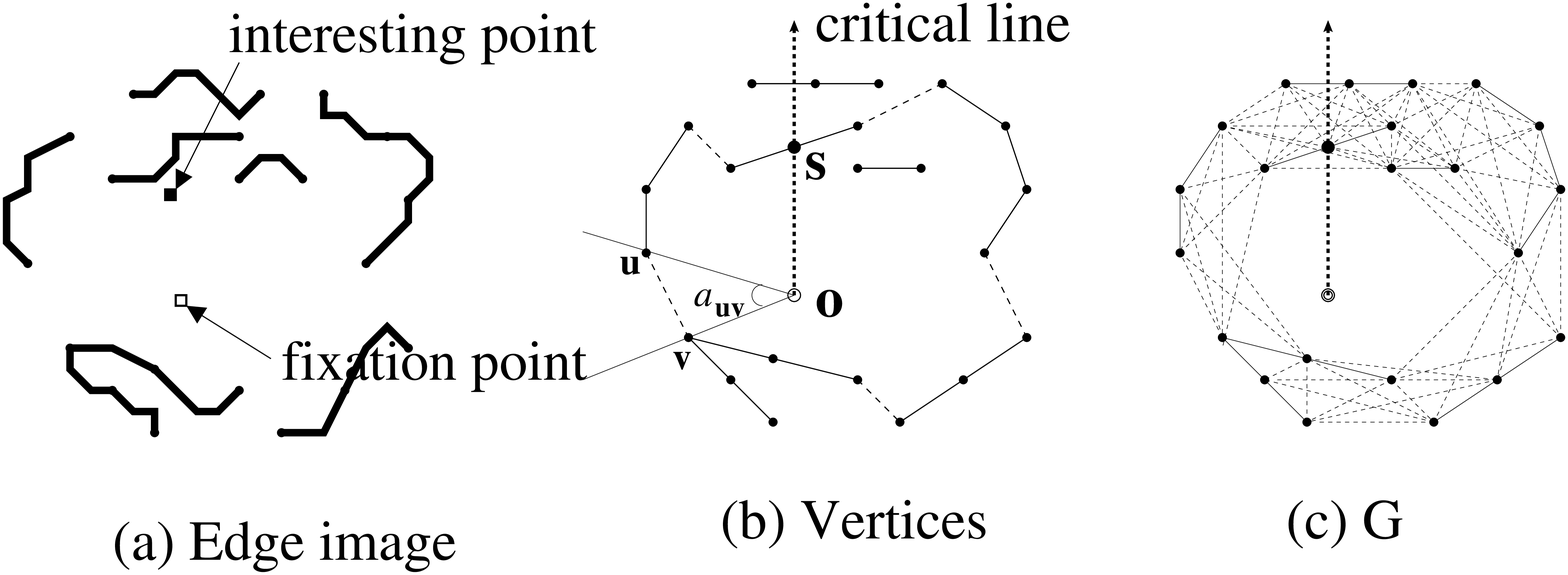}
\caption{Formulation of an angularly ordered graph from an edge image. (a)
is an edge image. Locations of a fixation point and and an interesting
point are shown. (b) shows a set of vertices. A pair of vertices from
the same contour fragment is connected with a solid line. The shortest $\theta$
enclosing cycle with $\theta>\alpha_{\textbf{uv}}$ is shown with
dashed and solid lines. (c) shows an angularly ordered graph with
$\theta=\pi/2$. A set of arcs with visual angles not less than $\theta$ are removed.}
\label{fig:GraphFormulation}
\end{figure}

See Figure \ref{fig:GraphFormulation} for an illustration of
these definition. In (a), an edge image with hypothetical
fixation and interesting points are shown. In (b), a set of
vertices obtained from the edge image of (a) are shown after
splitting of long contour fragments. Vertices from the same
contour fragment are shown connected by a solid line. Let
$\alpha_{\textbf{uv}}$ be the visual angle of the gap between
\textbf{u} and \textbf{v} seen from the fixation point. When
$\theta\ge\alpha_{\textbf{uv}}$, a shortest $\theta$-enclosing
cycle in (b) is the one delineated with dashed lines. The half
line starting from the fixation point and extending through the
start vertex plays an important role in our algorithms, and is
called \textit{critical line}. In (c), a corresponding
angularly ordered graph with $\theta=\pi/2$ is shown.

The problem is well-defined, as there are a finite number of
$\theta$-enclosing cycles in $G$ and each cycle has a finite
path distance measure. The algorithm of \cite{Kubota:POCV2010}
solves the problem when the shortest enclosing cycle consists
only of forward looking arcs.

\section{Algorithms}
In this section, we describe two algorithms that extend the
work of \cite{Kubota:POCV2010}.
\subsection{Algorithm I}
We transform $G$ into another graph $\hat{G}$ by taking the
following steps. First, we assign to each vertex in $G$ an
angle in $[0, 2\pi)$ determined by the visual direction of the
vertex from the fixation point. We use \textbf{s} as the
reference so that the angle of \textbf{s} is 0. We assume that
the angle increases in the counter-clockwise direction, but
this orientation is arbitrary. Thus, the angle of a vertex
slightly to the left (right) of \textbf{s} is close to 0
($2\pi$). Let $\angle{\textbf{u}}$ denote the angle of
\textbf{u}. Second, we add a target vertex, \textbf{t}, at the
same location with \textbf{s} but with
$\angle{\textbf{t}}=2\pi$ instead of 0. Third, we duplicate all
arcs incident on \textbf{s} and make them incident on
\textbf{t} instead of \textbf{s}. Finally, we remove arcs
between \textbf{u} and \textbf{v} if
$|\angle{\textbf{u}}-\angle{\textbf{v}}|\ge\theta$. Note that
arcs with $|\angle{\textbf{u}}-\angle{\textbf{v}}|\ge\theta$
are those that crosses the critical line where the angle jumps
from $2\pi$ to 0. All other arcs with
$|\angle{\textbf{u}}-\angle{\textbf{v}}|\ge\theta$ have been
removed at construction of $G$. Thus, this step eliminates all
arcs and only arcs that cross the critical line. Arcs removed
include those that are incident on \textbf{s} from the right
side of the critical line and those that are incident on
\textbf{t} from the left side of the critical line.

\begin{figure}
\centering
\includegraphics[width=2.5in]{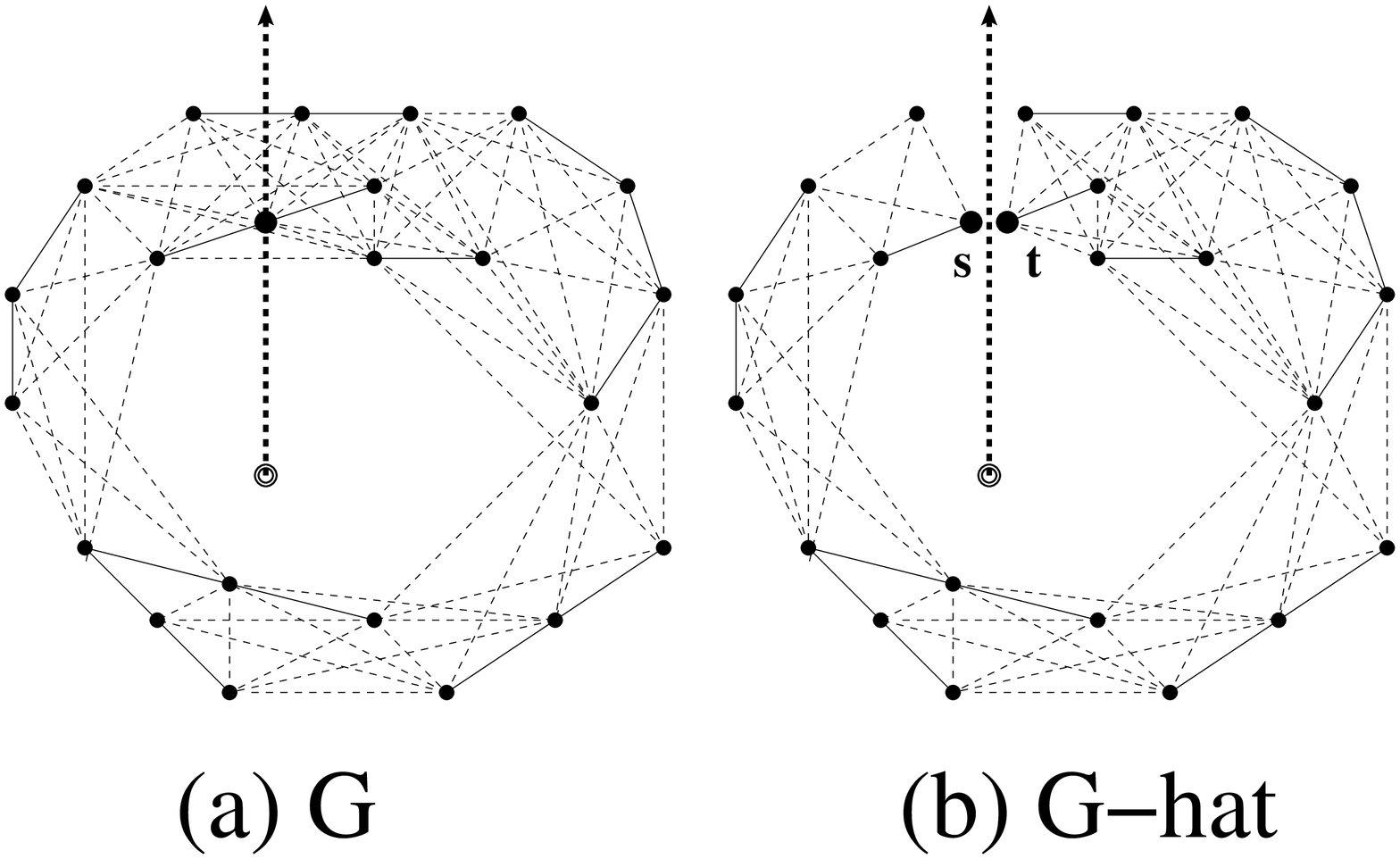}
\caption{(a) The same graph from Figure \ref{fig:GraphFormulation}(c).
(b) $\hat{G}$ with introduction of \textbf{t} and removal of arcs
crossing the critical line.}
\label{fig:GraphHatIllustration}
\end{figure}

We can visualize the entire steps as cutting $G$ at the
critical line as shown in Figure \ref{fig:GraphHatIllustration}
where (a) is the same G shown in Figure
\ref{fig:GraphFormulation}(c) and (b) is $\hat{G}$ derived from
it. It is clear from Figure \ref{fig:GraphHatIllustration}(b)
that every path from \textbf{s} to \textbf{t} is a $\theta$
enclosing one. We can find the shortest one among those by
applying a shortest path algorithm on $\hat{G}$. This is
basically Algorithm 1.

$\hat{G}$ may not have every $\theta$-enclosing cycle in $G$.
Missing ones are those that have arcs over the critical line.
Therefore, a shortest $\theta$-enclosing cycle in $G$ is a
shortest path in $\hat{G}$ if the cycle does not cross the
critical line more than once.

We can consider the algorithm of \cite{Kubota:POCV2010} as
finding a shortest path in $\hat{G}$ with undirected arcs
replaced by directed ones. In \cite{Kubota:POCV2010}, arcs are
incident only from a lower angle vertex to a higher or equal
angle one. In $\hat{G}$, the arcs are undirected, and we can
move in both directions to explore more options.

\begin{figure}
\centering
\includegraphics[width=3.in]{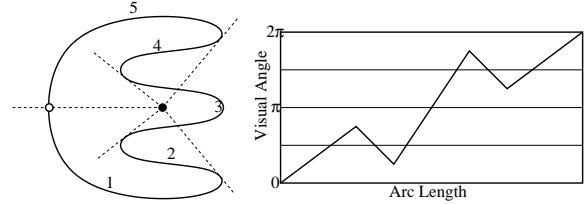}
\caption{A non-star shape with $K=5$. In the left, locations where the visual
angle changes the direction are marked by four tangent lines. In the right,
the same shape is graphed by its arc-length vs. the visual angle. Changes in the angular
direction appear changes in its slope in the graph.}
\label{fig:Eshape}
\end{figure}

We can apply Dijkstra algorithm to find a solution efficiently
in $O(|V|^2 \log|V|)$ where $|V|$ is the number of vertices in
$G$ (or $\hat{G}$). Another approach is to use the dynamic
programming idea of \cite{Kubota:POCV2010} but apply it in both
directions repeatedly. The advantage of the second approach is
that we can quickly extract a rough sketch of the underlying
object approximated by a star shaped contour in $O(|V|^2)$.
Subsequent processes in a processing chain can start as soon as
the approximate shape is extracted. We can think this approach
as Bellman-Ford algorithm with a specific visitation schedule
(i.e. repeated forward and backward directions in terms of the
visual angle). Thus, the algorithm still extracts the optimal
path in $\hat{G}$. More specifically, it will take $O(|V|^2K)$
to find the shortest path where $K-1$ is the number of changes
in the angular direction as we trace the path. A star shape has
$K=1$. The shape shown in Figure \ref{fig:Eshape} has $K=5$.

%We call this visitation schedule \textit{ angularly ordered
%iteration}.

\subsection{Algorithm II}

\begin{figure}
\centering
\includegraphics[width=4.in]{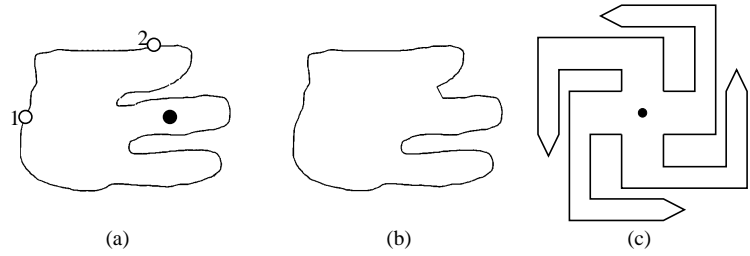}
\caption{(a) An example shape with a fixation and two starting points shown
in a solid disk and hollow disks, respectively. With the starting point
marked 1, Algorithm I will extract the complete figure. However, with the starting
point marked 2, Algorithm I will extract a partial shape shown in (b).
(c) A shape whose contour cannot be extracted by Algorithm I regardless of the
selection of a starting point. }
\label{fig:Algorithm1Illustration}
\end{figure}

Algorithm I effectively extends the search space for a
$\theta$-enclosing cycle from star-shaped ones to more general
ones. A restriction is that the cycle cannot cross the critical
line more than once. Thus, Algorithm I may fail to extract
highly articulate shapes. For example, see Figure
\ref{fig:Algorithm1Illustration}(a). A fixation point is shown
with a solid disk and two starting positions are shown with
hollow disks. When the starting position is the one marked 1,
the critical line only crosses the shape once, and Algorithm I
will be able to extract the whole shape. However, when it is
the one marked 2, the shape crosses the critical line three
times, and Algorithm I fails to extract the whole shape. The
result is shown in Figure \ref{fig:Algorithm1Illustration}(b).
A careful placement of the interesting point (and thus the
resulting start vertex) can usually circumvent the problem.
However, there exist shapes without such points. See Figure
\ref{fig:Algorithm1Illustration}(c). Although this example
looks contrived and no such shape is found in nature, our goal
is to automate the starting point selection process and it will
be difficult to make the process responsible for avoiding the
shape from crossing the critical line. This subsection
describes our second algorithm, which extends the search space
and circumvents the limitation of Algorithm I to some extent.

\begin{figure}
\centering
\includegraphics[width=4in]{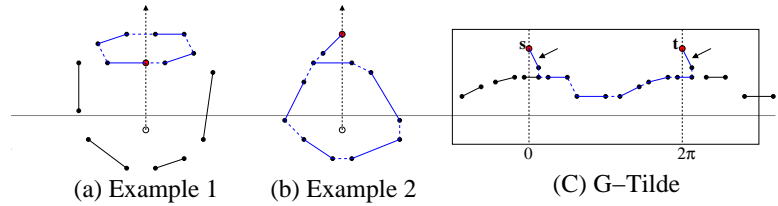}
\caption{(a) An example shape. By allowing arcs crossing
the critical line, a non-enclosing cycle as shown in blue
can be admitted. (b) Another example shape. By replicating vertices circularly,
a non-simple boundary can be admitted as shown in blue here. (c)
However, the path is actually simple in $\tilde{G}$. Two arrows points to
arcs that correspond to the same contour fragment in (b).
}
\label{fig:FlawedApproaches}
\end{figure}

A quick investigation may bring two ways to extend Algorithm I.
One is to keep arcs whose visual angles are less than $\theta$
regardless of them crossing the critical line or not. The
approach is flawed as the search space include non-$\theta$
enclosing paths from \textbf{s} to \textbf{t}. See Figure
\ref{fig:FlawedApproaches} (a) where a set of blue contours
forms a shortest cycle that is not $\theta$-enclosing one. The
other approach is to extend $\hat{G}$ by circular replication
of vertices. Denote the graph $\tilde{G}$. In $\tilde{G}$, all
paths from \textbf{s} to \textbf{t} are $\theta$ enclosing
ones. Along the path, the visual angle can go negative or over
$2\pi$. Such instances accommodate arcs crossing the critical
line. However, the approach is also flawed, as $\tilde{G}$
permits a path that visits the same contour fragment more than
once, although the path in $\tilde{G}$ is simple. See Figure
\ref{fig:FlawedApproaches} (b) and (c). The former shows a
shape represented by extracted vertices. The latter shows
$\tilde{G}$ derived from the shape of (b) in polar coordinate.
A blue colored path in (c) is the shortest one from \textbf{s}
to \textbf{t} and corresponds to the blue boundary shown in
(b). Two arcs pointed by two arrows in (c) correspond to the
same contour fragment in (b).

As observed above, we cannot simply apply a shortest path
algorithm to $\tilde{G}$ as it may result in a non-simple
boundary. Instead, we use a greedy way to improve the solution
of Algorithm I. A new algorithm (Algorithm II) first runs
Algorithm I to find an optimum path in $\hat{G}$. It then
extends the graph from $\hat{G}$ to $\tilde{G}$, and replaces
the existing path from \textbf{s} to \textbf{t} when a better
alternative is found. Thus, the approach does not guarantee the
shortest cycle in $G$ but does guarantee that the solution is
not worse than that of Algorithm I.

Before describing details, some definitions and notations are
in order. We use $P$ to denote the current path from \textbf{s}
to \textbf{t}. We call $\textbf{u}\in
V\setminus\left\{\textbf{s},\textbf{t}\right\}$ a replica of
\textbf{v} if $\textbf{u}\ne\textbf{v}$ but corresponds to the
same contour end point of \textbf{v}. We call \textbf{u}
\textit{dormant} if its replica is currently in $P$. The
\textit{dormant set} of $\tilde{G}$ is a set of dormant
vertices. In other words, a vertex \textbf{u} is in the dormant
set if $\textbf{u}\notin P$ and one of its replica is in $P$.
We denote the dormant set $D$. A path from \textbf{u} to
\textbf{v} is \textit{consistent} if no pair of vertices on the
path are replica of each other. Exclusion of \textbf{s} and
\textbf{t} from the replica definition is a minor technical one
as they do correspond to the same end-point and thus making $P$
inconsistent.

Roughly speaking, Algorithm II extends the shortest path tree
found by Algorithm I to one in $\tilde{G}$ with exclusion of
all vertices in the dormant set. When we find a shorter path
from \textbf{s} to \textbf{t}, we check if it is consistent. If
so, we replace the existing path by the better alternative. The
consistency is the absolute requirement for the solution to be
simple in the image space. So why do we want to consider both
dormancy and consistency? We use dormancy to prevent many
inconsistent branches from forming. This will help growing
consistent branches to reach $P$.

The Algorithm II is shown below. $d(\textbf{u})$ is the current
path distance of \textbf{u} from \textbf{s}.
$w(\textbf{u},\textbf{v})$ is the weight of an arc
$(\textbf{u},\textbf{v})$. $\textbf{v}_\pi$ is the parent of
$\textbf{v}$ in the tree. Line 8 checks if the update will
alter the current $P$. However, we allow it only if
$\textbf{s}\rightarrow\textbf{u}$ is consistent (Lines 9 and
10). Since dormant vertices are excluded,
$\textbf{s}\rightarrow\textbf{u}$ being consistent implies
$\textbf{s}\rightarrow\textbf{u}\rightarrow\textbf{v}\rightarrow\textbf{t}$
being consistent. When the dormant set is updated in Line 11,
we need to set $d$ of all descendants under each vertex in the
new dormant set to $\infty$ to prevent any illegal path from
forming from the descendants.

\begin{algorithm}[h]\label{algo:AlgorithmII}
%\SetLine
\KwIn{$\hat{G}$, $\textbf{s}$, $\textbf{t}$}
\KwOut{$\textbf{s}\rightarrow\textbf{t}$:a path from \textbf{s} to \textbf{t}}
Apply Algorithm I on $\hat{G}$\\
Extend $\hat{G}$ to $\tilde{G}$.\\
Find $D$, a dormant set of vertices given $P=\textbf{s}\rightarrow\textbf{t}$\\
\Repeat{there is no change} {
    \ForEach{$(\textbf{u},\textbf{v})$ in $\tilde{G}$}
    {
       \If{$\textbf{u}\notin D$ and $\textbf{v}\notin D$}
       {
            \If{$d(\textbf{u}) + w(\textbf{u},\textbf{v})<d(\textbf{v})$}
            {
                 \If{$\textbf{u}\notin P$ and $\textbf{v}\in P$}
                 {
                     \If{$\textbf{s}\rightarrow\textbf{u}$ is NOT consistent}
                     {
                         continue;
                     }
                     Update $D$ given $P=\textbf{s}\rightarrow\textbf{u}\rightarrow\textbf{v}\rightarrow\textbf{t}$;\\
                 }
                 $\textbf{v}_\pi=\textbf{u}$;\\
                 $d(\textbf{v})=d(\textbf{u}) + w(\textbf{u},\textbf{v})$;\\
            }
       }
    }
}
\caption{Algorithm II}
\end{algorithm}

Note that the algorithm is guaranteed to terminate since there
can be only a finite number of ways the path cost
$P=\textbf{s}\rightarrow\textbf{t}$ can be reduced, and within
a fixed $P$, the algorithm is the Bellman-Ford, which is
guaranteed to terminate in $O(|V||E|)$. The possible number of
updates is upper bounded by $O(2^{|V|})$ although the actual
number is much smaller. The maintenance of the dormant set
takes $O(|V|)$ with a tree data structure to maintain the
shortest path tree and does not contribute to the overall
complexity.

\begin{figure}
\centering
\includegraphics[width=2.5in]{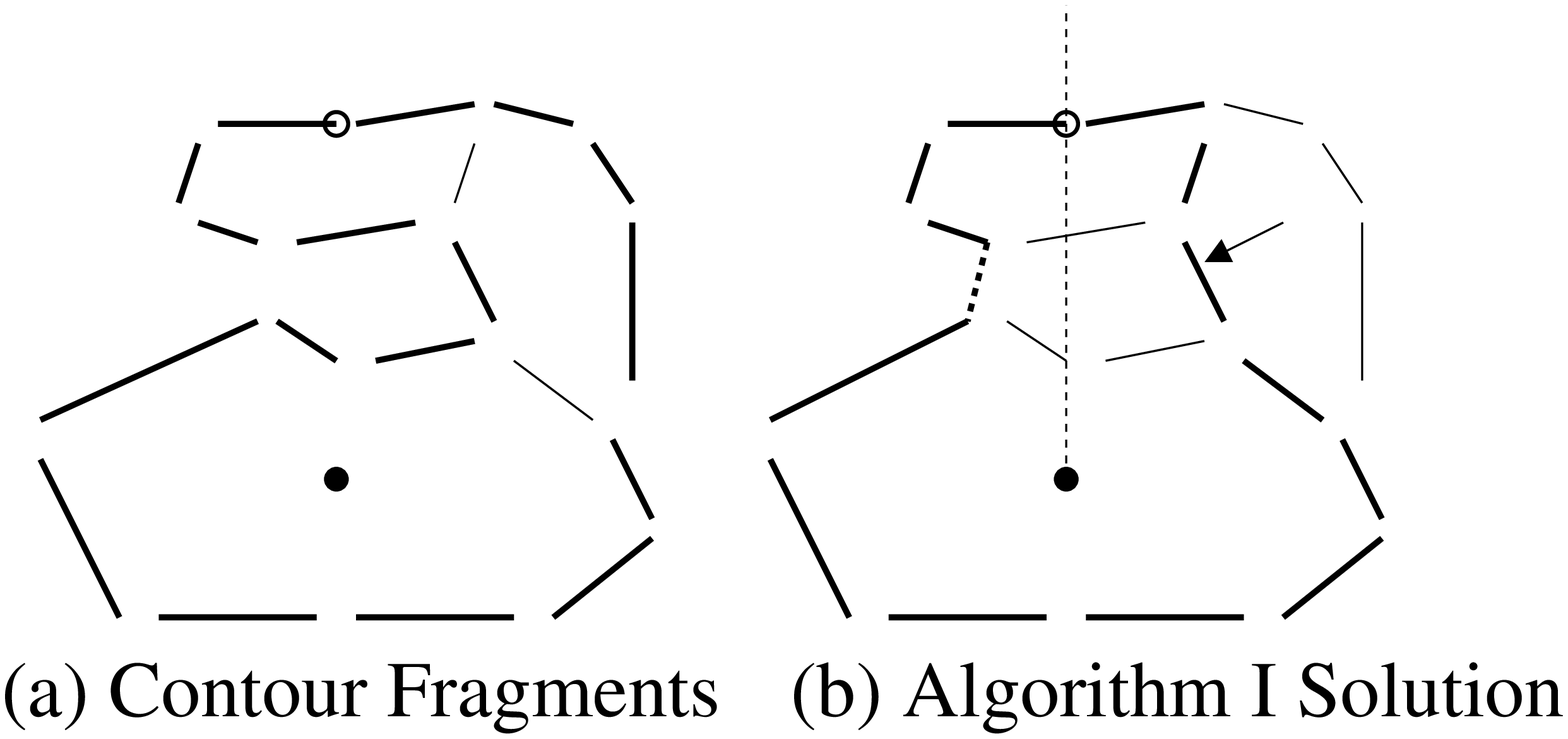}
\caption{(a) An example graph. Thick lines shows the optimal
$\theta$ enclosing cycle. (b) The solution after Algorithm I.
A contour fragment pointed by the arrow has the angle in
$[3\pi/2,2\pi]$ while the optimal configuration of (a) has the same
fragment in $[-\pi/2,0]$.}
\label{fig:AlgorithmIINonOptimal}
\end{figure}

Since the algorithm is a greedy one, it may only find a locally
optimum one. This can happen when the result of Algorithm I
uses a replica of a vertex in an optimal path. For a such
example, see Figure \ref{fig:AlgorithmIINonOptimal} where (a)
shows contour fragments with thick ones delineating the optimum
solution, and (b) shows the result of Algorithm I. In (b), the
fragment pointed by the arrow has the visual angle somewhere
between $3\pi/2$ and $2\pi$ while the same fragment appears in
(a) as somewhere between $-\pi/2$ and 0. This means that a
vertex of the fragment in (b) (call it \textbf{x}) is a replica
of that in (a) (call it \textbf{y}). While \textbf{x} is in
$P$, \textbf{y} remains dormant. Thus, Algorithm II cannot use
it to improve the current solution. It first needs to remove
\textbf{x} from $P$ so that \textbf{y} is removed from the
dormant set. However, the step will likely to increase the path
cost.

\section{Experiments}
We implemented three algorithms: the algorithm of
\cite{Kubota:POCV2010}, Algorithm I, and Algorithm II. The
implementation of \cite{Kubota:POCV2010} is done by replacing
undirected arcs in $\hat{G}$ with directed ones that are
forward looking.

The algorithms were tested on 692 pairs of a fixation point and
a starting point, placed manually, on 520 images in which 300
were taken from the Berkeley Segmentation Dataset and 220 were
collected from other sources. For each image, Canny edge
detector in MATLAB with the default setting is applied,
connected edgels are sequenced, and those contour fragments
that are less than 5 pixels are removed. A set of contour
fragments obtained by these steps, a fixation point, and a
starting point are the inputs to the algorithms. We used $L=10$
and $\theta=\pi/2$. The number of vertices in the resulting
graph ranged from 134 to 2666 with the average of 1026.

Table \ref{table:Performance} summarizes the performance in
terms of the path cost ($d$) and computation time in seconds
($t$) for each algorithm. The computation time does not include
time it took for Canny edge detection and sequencing of edgels.
The data are collected on a PC with a 2.67GHz Intel Core i7 CPU
with 4GB of memory. The algorithms were written in C++ and
built with Visual Studio 2010.

The method of \cite{Kubota:POCV2010} always has the largest $d$
and the smallest $t$, Algorithm I always has the second largest
$d$ and the second largest $t$, and Algorithm II always has the
smallest $d$ and the largest $t$. In average, Algorithm I took
about $50\%$ additional time than the method of
\cite{Kubota:POCV2010} and Algorithm II took about 6 times more
than Algorithm I. In the experiment, $\tilde{G}$ is expanded to
$-2\pi$ to $4\pi$. We could reduce the computational time of
Algorithm II somewhat by reducing it to $-\pi$ to $3\pi$
without hampering the performance.

\begin{table}
\centering \label{table:Performance} \caption{Comparisons of
three algorithms}
\begin{tabular}{|c|c|c|c|c|c|c|}
  \hline
  Method & Max $d$ & Mean $d$ & Min $d$ & Max $t$ & Mean $t$ & Min $t$ \\
  \hline
  \cite{Kubota:POCV2010} & 4295 & 268 & 15 & 0.83 & 0.13 & 0.003 \\
  Algorithm I & 4198 & 184 & 13.8 & 1.65 & 0.23 & 0.004 \\
  Algorithm II & 4163 & 171 & 13.8 & 8.71 & 1.28 & 0.011 \\
  \hline
\end{tabular}
\end{table}

\begin{figure}
%\centering
%\includegraphics[width=2.5in]{26031B.eps}
\includegraphics[width=2.5in]{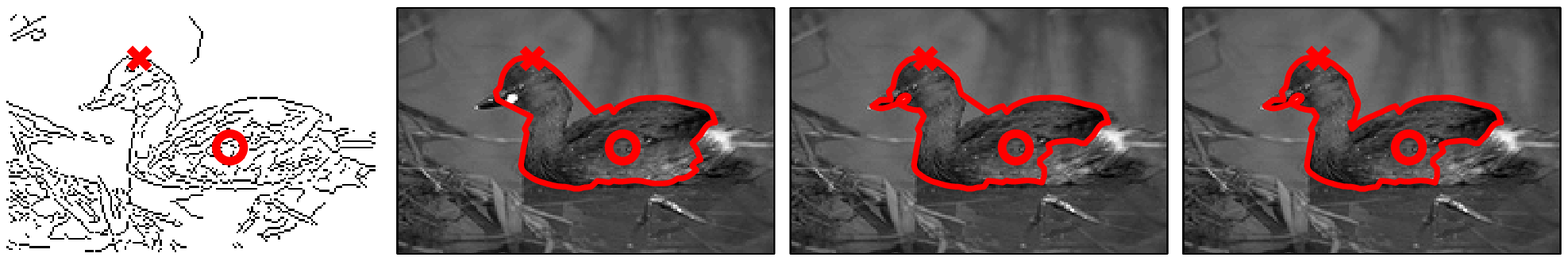}
\includegraphics[width=2.5in]{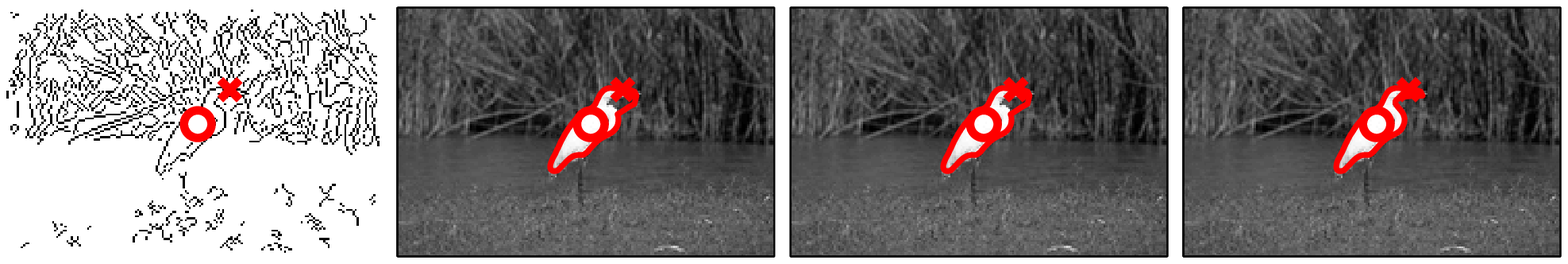}
\includegraphics[width=2.5in]{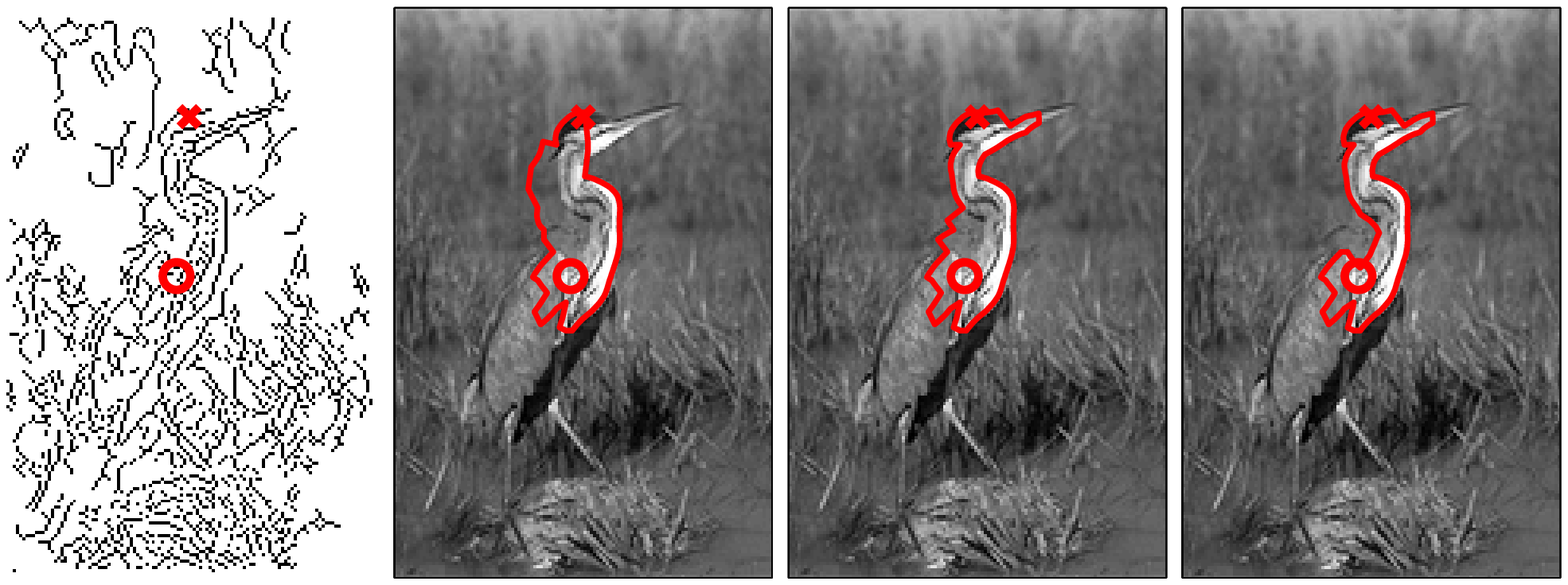}
\includegraphics[width=2.5in]{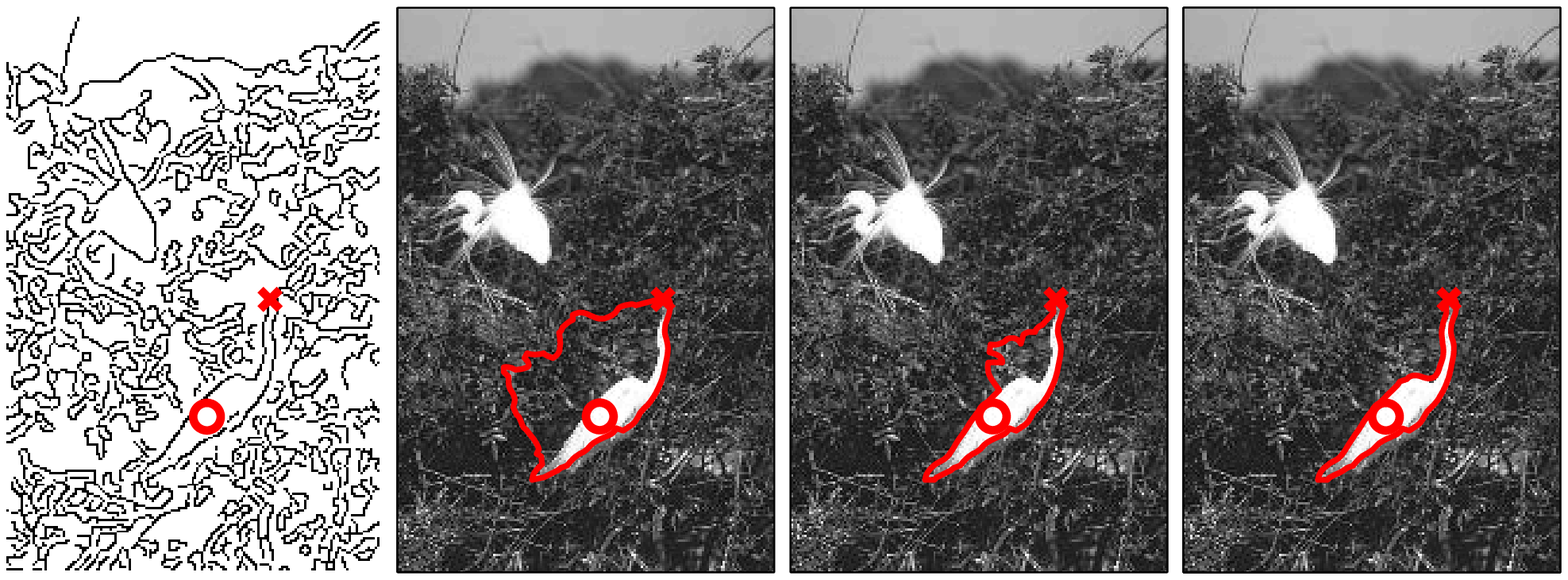}
\includegraphics[width=2.5in]{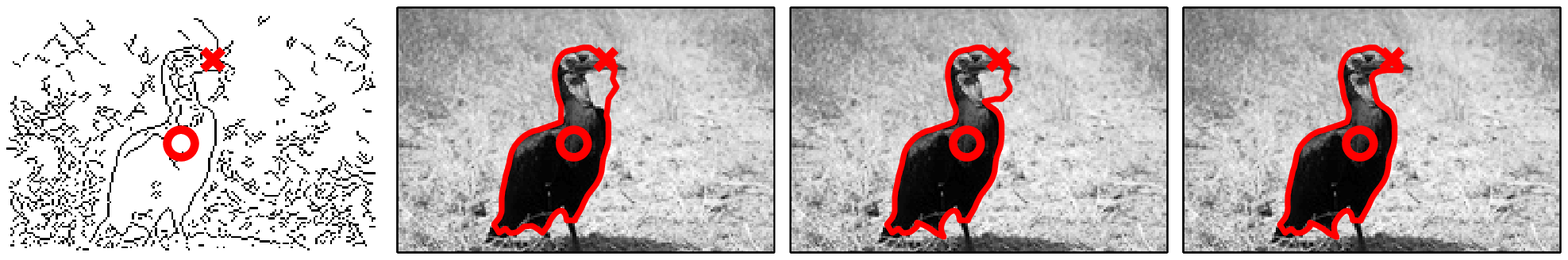}
\includegraphics[width=2.5in]{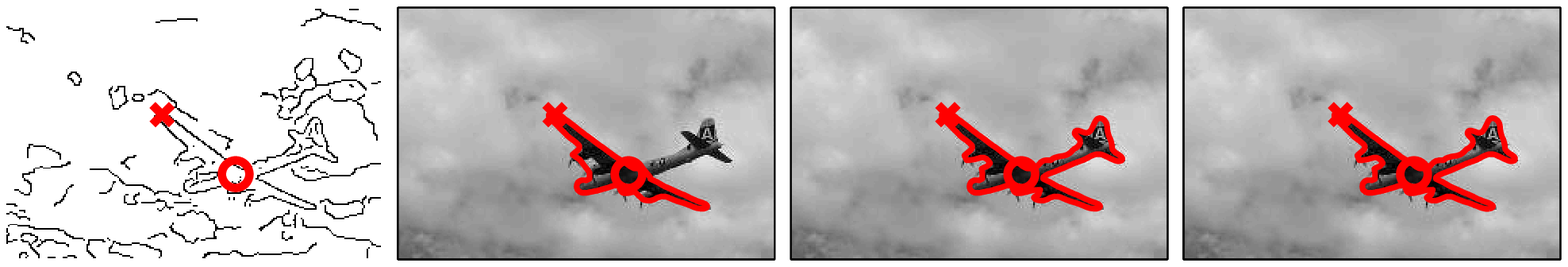}
\includegraphics[width=2.5in]{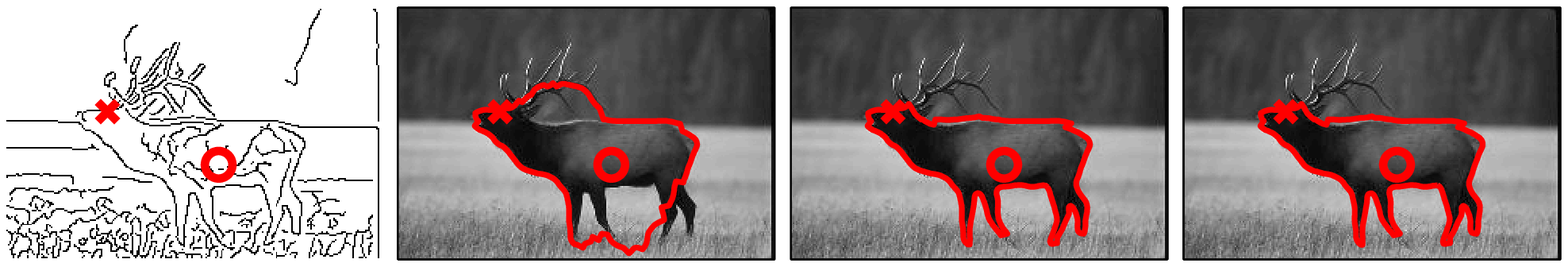}
\includegraphics[width=2.5in]{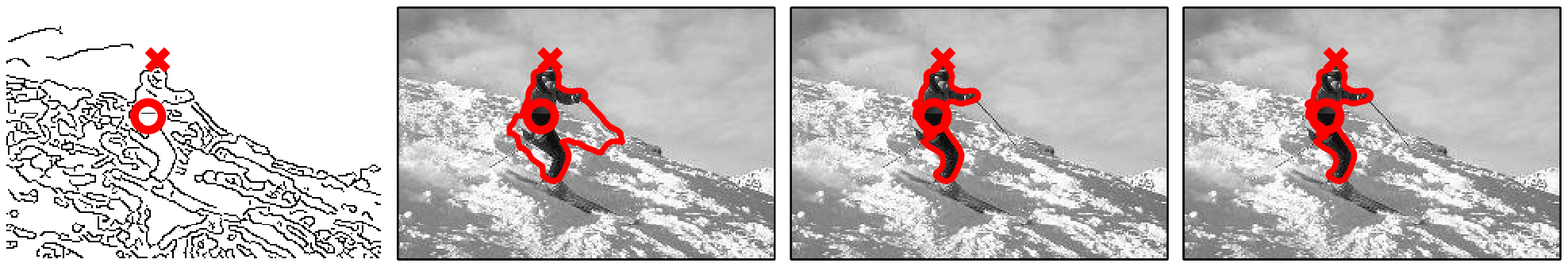}
\includegraphics[width=2.5in]{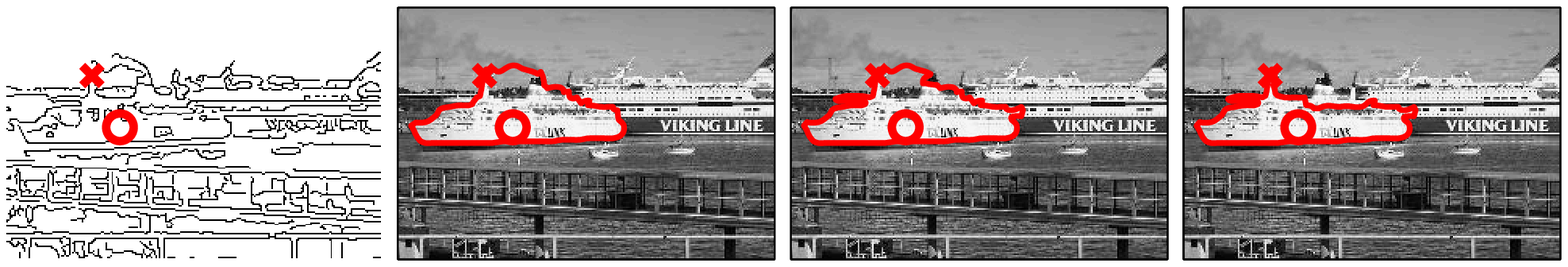}
\includegraphics[width=2.5in]{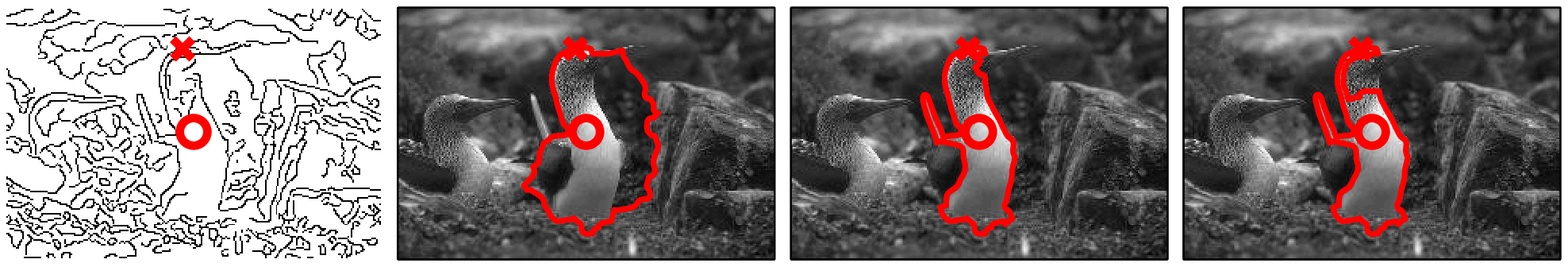}
\includegraphics[width=2.5in]{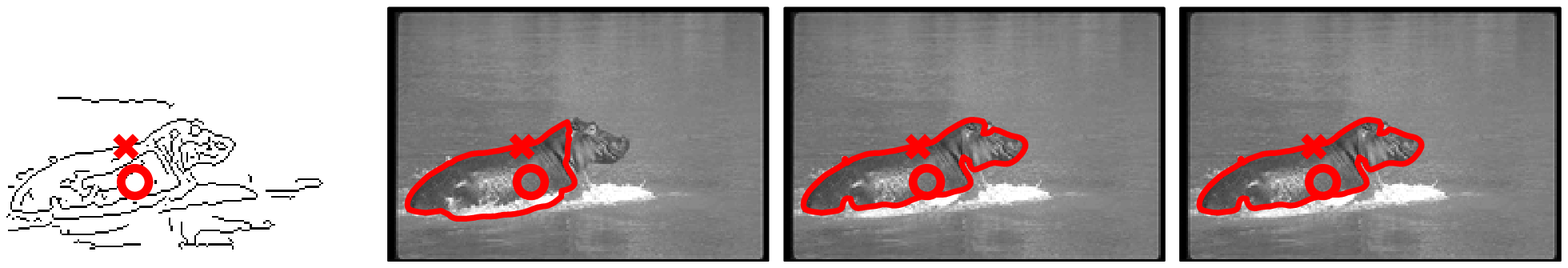}
\includegraphics[width=2.5in]{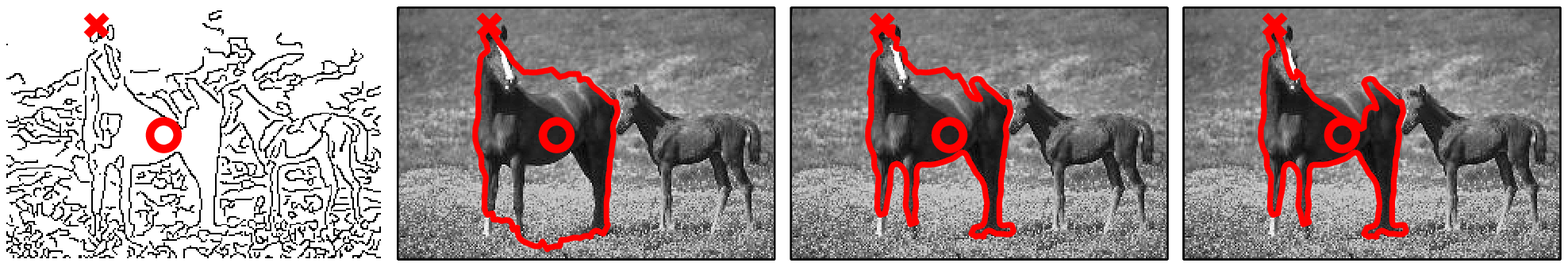}
\includegraphics[width=2.5in]{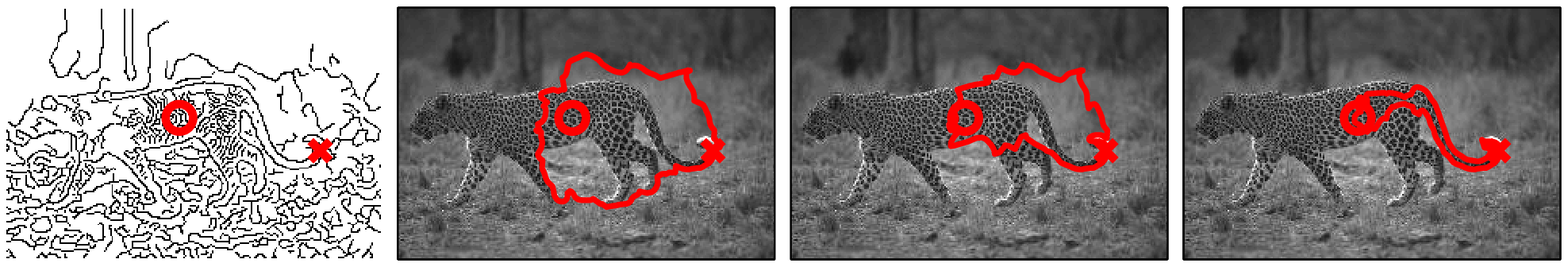}
\includegraphics[width=2.5in]{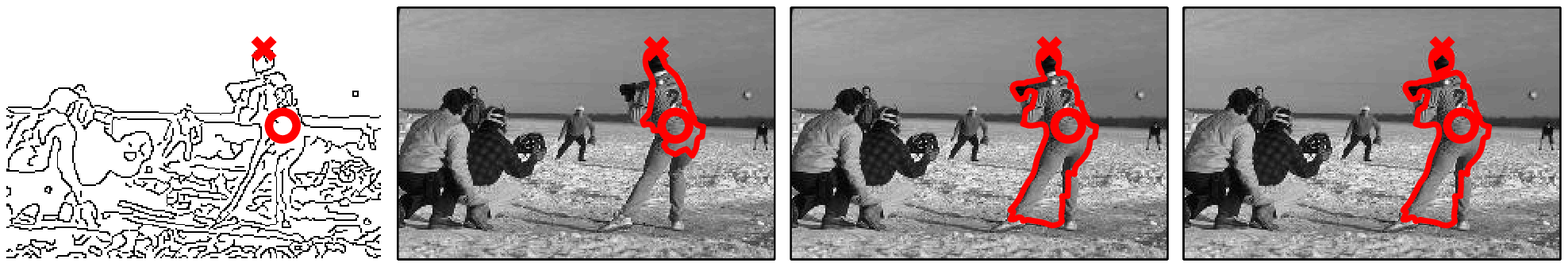}
\includegraphics[width=2.5in]{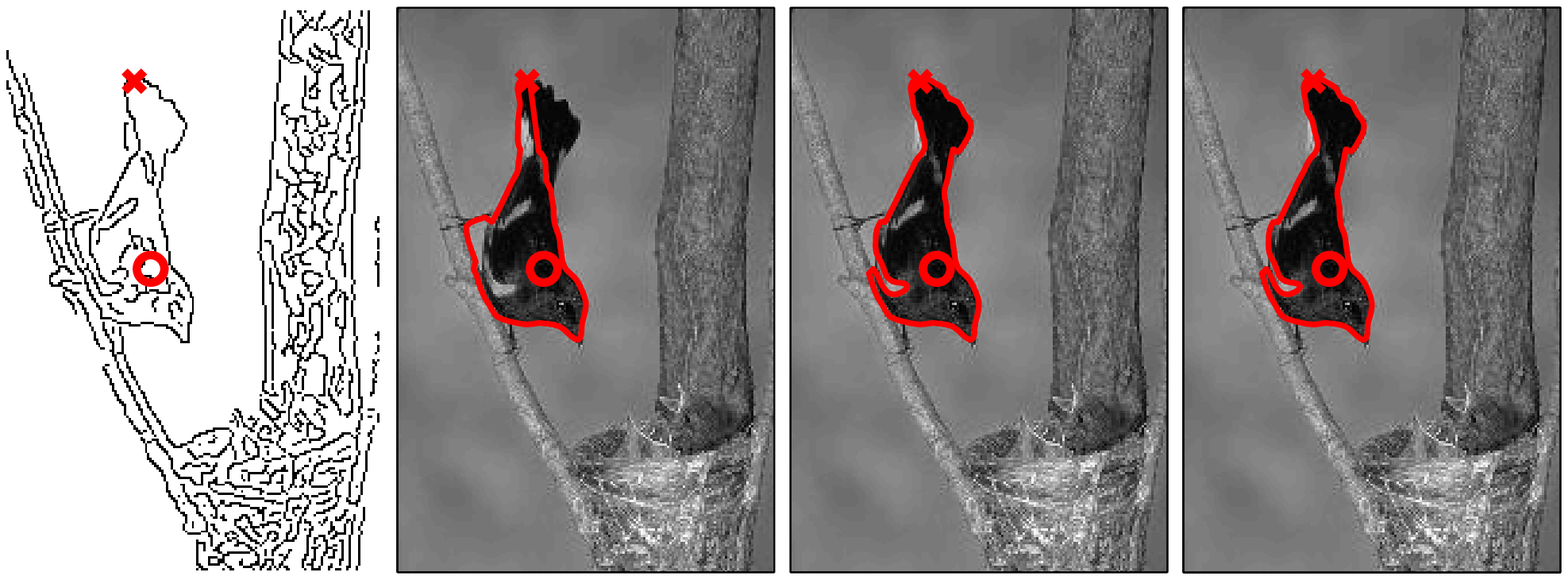}
\includegraphics[width=2.5in]{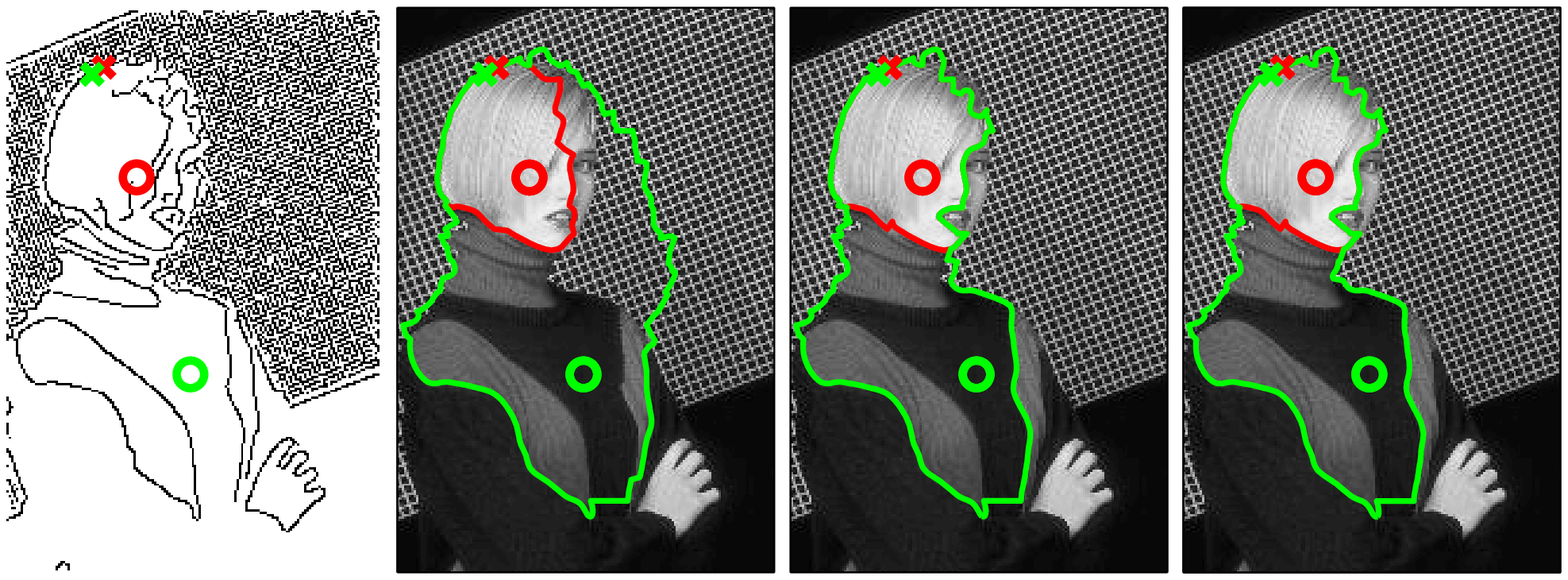}
\includegraphics[width=2.5in]{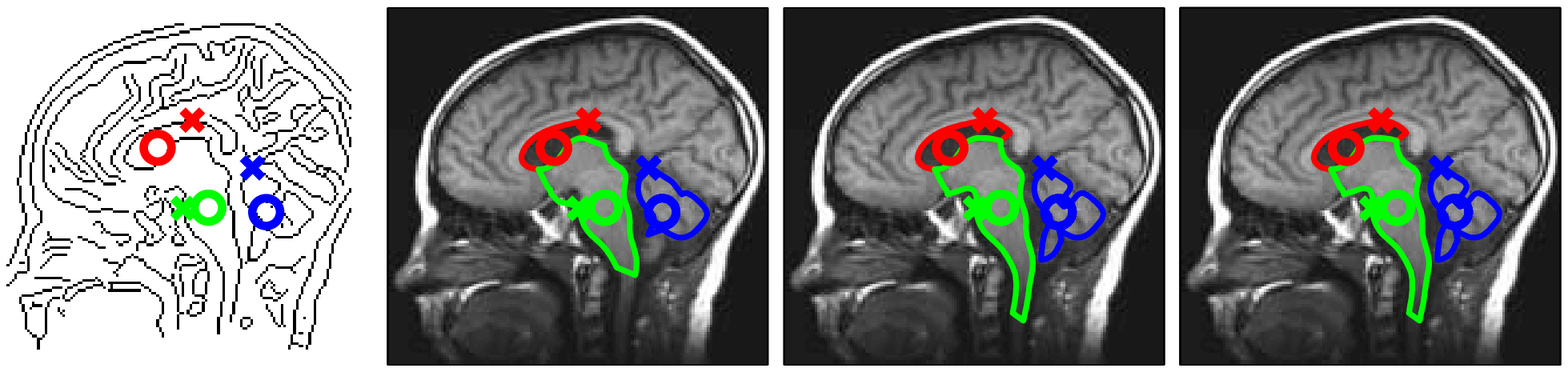}
\includegraphics[width=2.5in]{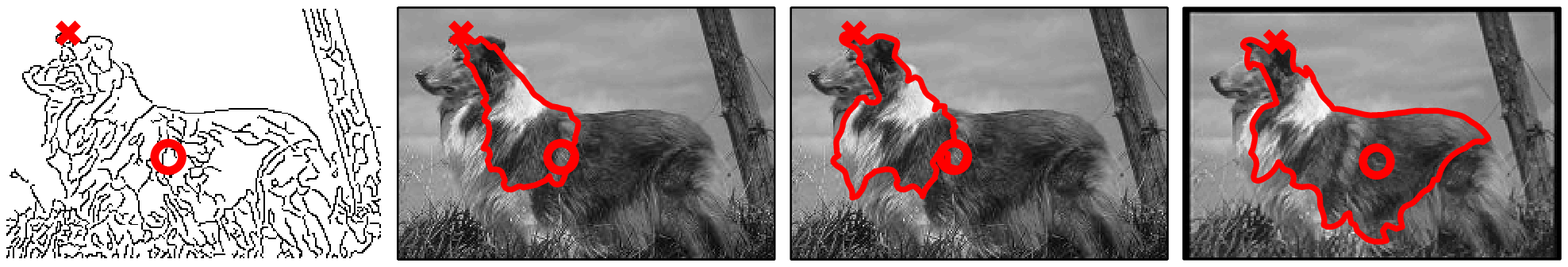}
\includegraphics[width=2.5in]{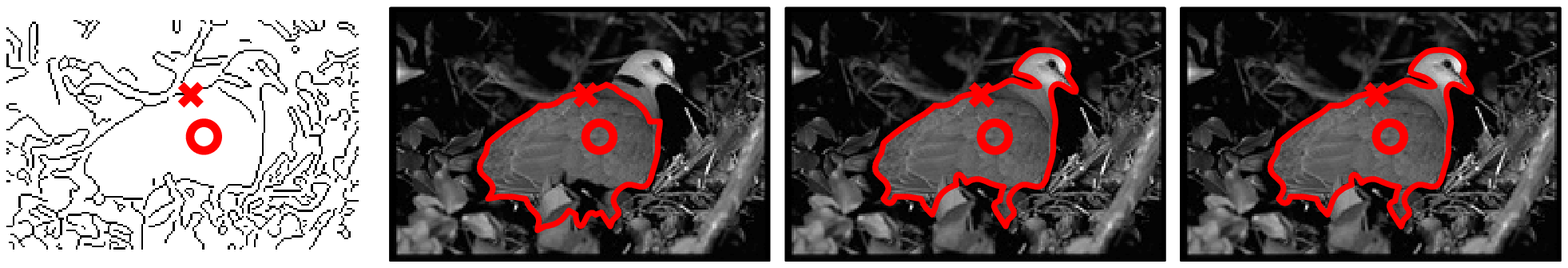}
\includegraphics[width=2.5in]{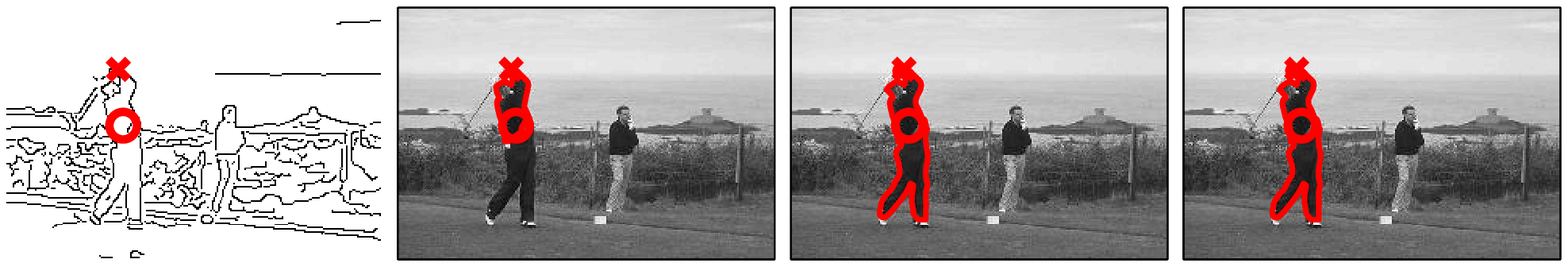}
\includegraphics[width=2.5in]{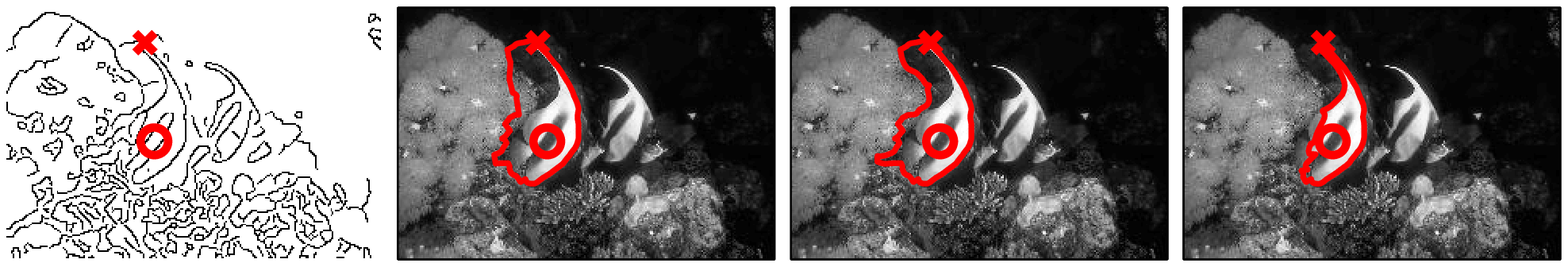}
\includegraphics[width=2.5in]{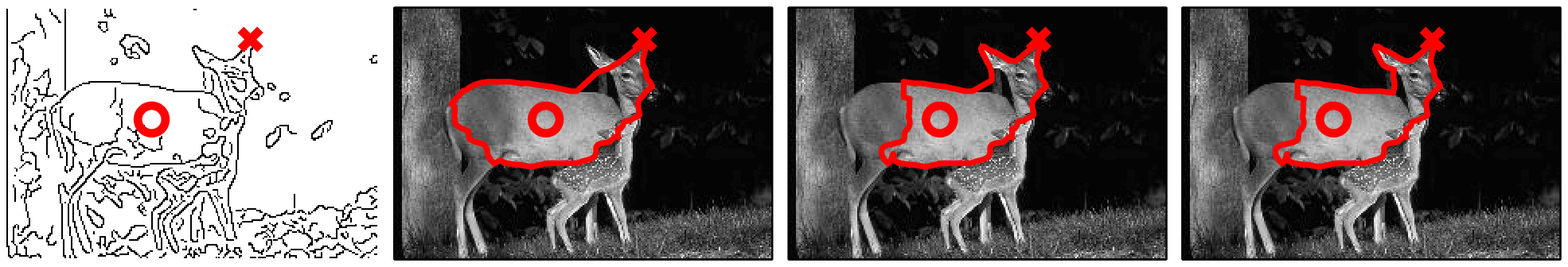}
\includegraphics[width=2.5in]{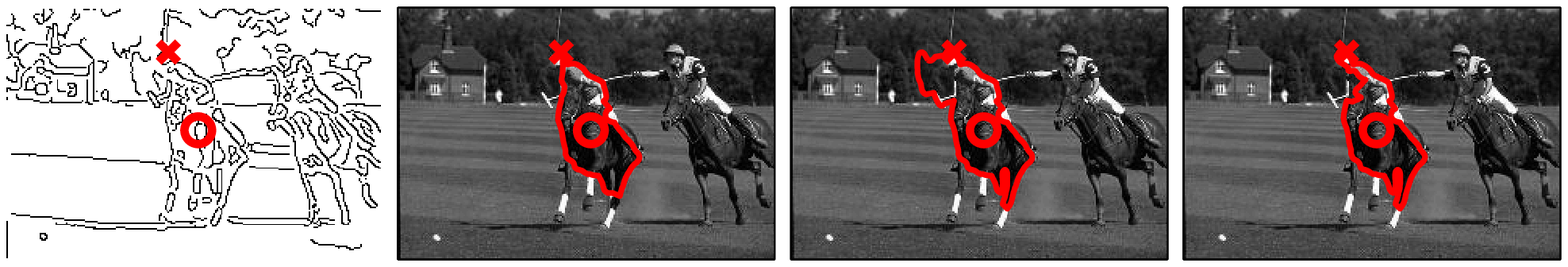}
\includegraphics[width=2.5in]{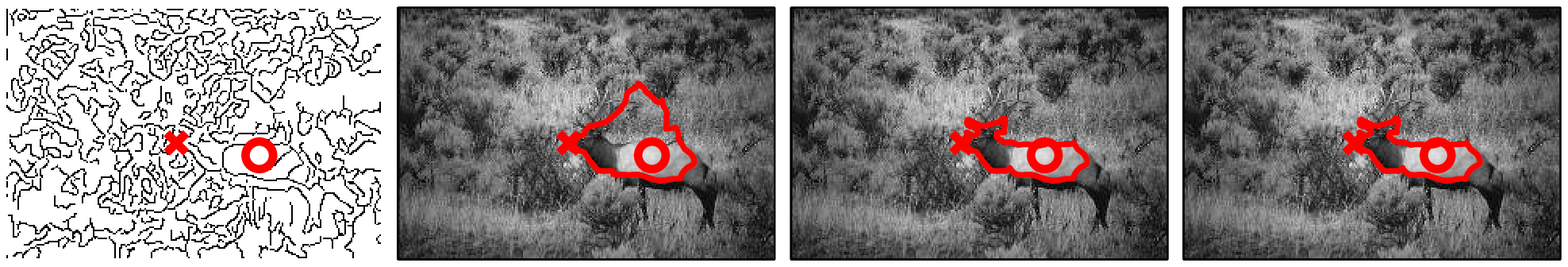}
\includegraphics[width=2.5in]{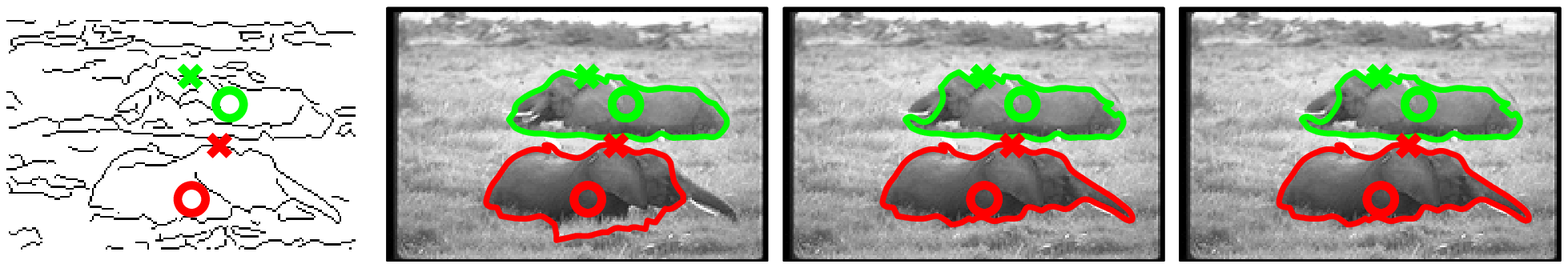}
\includegraphics[width=2.5in]{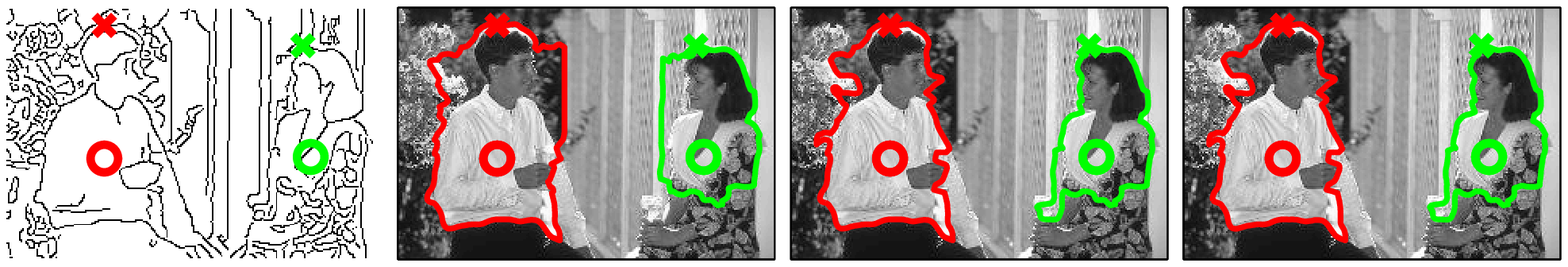}
\includegraphics[width=2.5in]{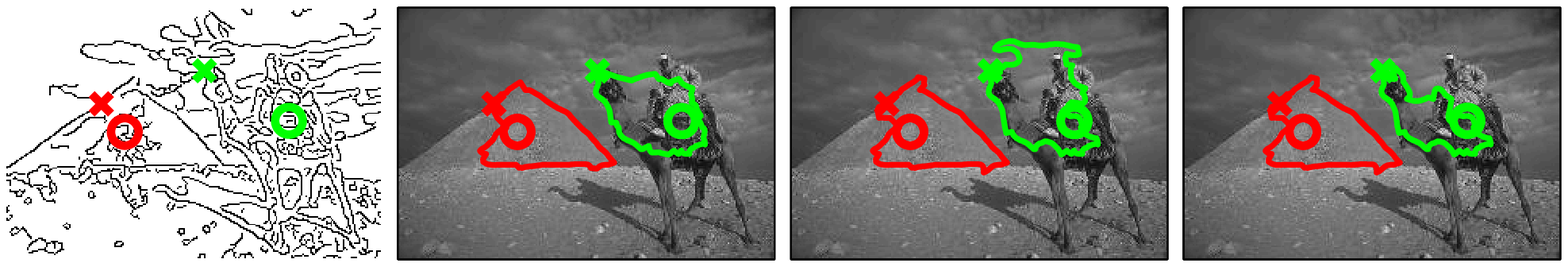}
\includegraphics[width=2.5in]{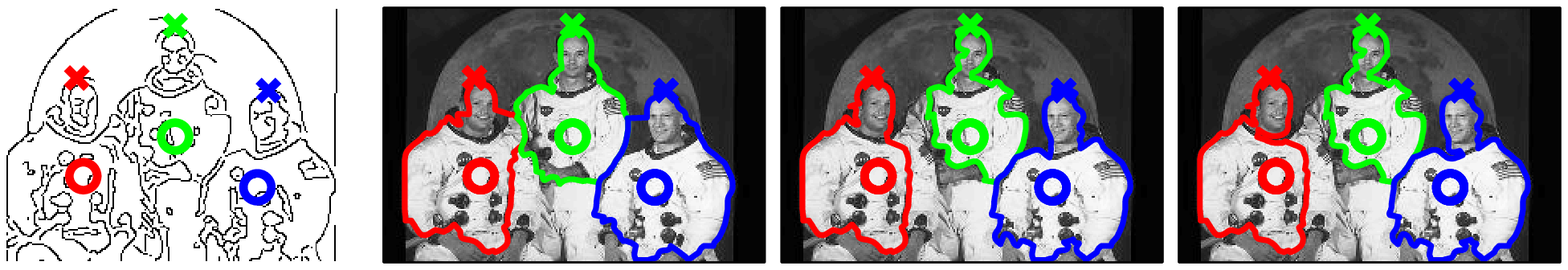}
\caption{Contour integration results. Shown are, from left to right,
a Canny edge image, the result of \cite{Kubota:POCV2010}, the result of
Algorithm I, and the result of Algorithm II. In each image, a hollow
circle is the fixation point and a cross mark is the starting point. }
\label{fig:Results}
\end{figure}

Figure \ref{fig:Results} shows some results from the
experiment. A small circle identifies the fixation point, a
small cross identifies the starting point, and a contour shows
the result of the corresponding algorithm. They are color coded
so that multiple results can appear when multiple sets of
fixation and start points are provided on the same image.
Algorithm II tends to delineate a tighter boundary than
Algorithm I and Algorithm I in turn tends to delineate a
tighter boundary than that of \cite{Kubota:POCV2010}.

\section{Conclusion}
The paper introduced two algorithms that improved the
performance of \cite{Kubota:POCV2010}. In addition to an edge
image, the algorithms take a fixation point and an interesting
point as inputs, which can drive the segmentation toward a
particular area and a particular scale. The underlying premise
is that providing these two points is simpler than providing a
bias\cite{Yu:NIPS2001} or suppressing more plausible
solutions\cite{Wang:RatioContour}\cite{Mahamud:PO}. By
providing multiple pairs of fixation and interesting points,
the algorithms can decompose the image into a meaningful set of
possibly overlapping figures. To automate this point generation
process may be more feasible than fully automated segmentation.

As in \cite{Kubota:POCV2010}, we consider edge information only
and very simplistic formulation of arc weights, namely the
squared distance between fragment end points. The purpose for
the simplicity is to keep our focus on development of shortest
path graph algorithms for the stated problem. Due to this
simplicity, results of the algorithms can have jagged
appearance. In the near future, we will incorporate color
information and geometrical information (continuity and
curvature, for example) of the fragments to improve
applicability of the algorithm to natural images.

Another issue for future study is automated placement of a
fixation point and an interesting point, as mentioned above.
Various studies have been conducted on these issues
\cite{Aanas:ICCV2012}\cite{Itti:PAMI1998}\cite{Koostra:CognComp2011}
and we will investigate applicability of the works to our
framework.

%\section*{Acknowledgement}
%This work is supported by NSF grant and Susquehanna University
%Faculty Research grant.
\bibliography{../bib/all}
\bibliographystyle{splncs}

\end{document}